\documentclass[lettersize,journal]{IEEEtran}
\usepackage{amsmath,amsfonts}
\usepackage{algorithmic}
\usepackage{algorithm}
\usepackage{array}
\usepackage[caption=false,font=normalsize,labelfont=sf,textfont=sf]{subfig}
\usepackage{textcomp}
\usepackage{stfloats}
\usepackage{url}
\usepackage{verbatim}
\usepackage{graphicx}
\usepackage{cite}
\usepackage{hyperref}       % hyperlinks
\usepackage{booktabs}       % professional-quality tables
\usepackage{nicefrac}       % compact symbols for 1/2, etc.
\usepackage{microtype}      % microtypography
\usepackage{xcolor}         % colors

\usepackage{xspace}
\usepackage{float}
\usepackage{mathtools}
\usepackage{multirow}
\usepackage{makecell}

\usepackage[edges]{forest}
\usepackage{tikz}
\usetikzlibrary{trees}
\tikzstyle{every node}=[draw=black,thick,anchor=west,font=\footnotesize,minimum height=2em]

\newtheorem{definition}{Definition}

\makeatletter
\DeclareRobustCommand\onedot{\futurelet\@let@token\@onedot}
\def\@onedot{\ifx\@let@token.\else.\null\fi\xspace}
 
\def\eg{\emph{e.g}\onedot} 
\def\ie{\emph{i.e}\onedot}

\def\etal{\emph{et al}\onedot}
\makeatother
\begin{document}

\title{Data-centric Graph Learning: A Survey}

\author{Yuxin Guo*, Deyu Bo*\thanks{\IEEEauthorrefmark{1} The first two authors contributed equally to this work.}, Cheng Yang$^\dag$, Zhiyuan Lu, Zhongjian Zhang, Jixi Liu, Yufei Peng,
Chuan Shi$^\dag$\thanks{\IEEEauthorrefmark{2} Corresponding authors: \{yangcheng,shichuan\}@bupt.edu.cn}% <-this % stops a space
\IEEEcompsocitemizethanks{\IEEEcompsocthanksitem School of Computer Science, Beijing University of Posts and Telecommunications, Beijing, China, 100876.}}

\markboth{Journal of \LaTeX\ Class Files,~Vol.~14, No.~8, August~2021}%
{Shell \MakeLowercase{\textit{et al.}}: A Sample Article Using IEEEtran.cls for IEEE Journals}

\IEEEpubid{0000--0000/00\$00.00~\copyright~2021 IEEE}

\maketitle

\begin{abstract}
The history of artificial intelligence (AI) has witnessed the significant impact of high-quality data on various deep learning models, such as ImageNet for AlexNet and ResNet. 
Recently, instead of designing more complex neural architectures as model-centric approaches, the attention of AI community has shifted to data-centric ones, which focuses on better processing data to strengthen the ability of neural models.
Graph learning, which operates on ubiquitous topological data, also plays an important role in the era of deep learning.  
In this survey, we comprehensively review graph learning approaches from the data-centric perspective, and aim to answer three crucial questions: \textit{(1) when to modify graph data}, \textit{(2) what part of the graph data needs modification} to unlock the potential of various graph models, and \textit{(3) how to safeguard graph models} from problematic data influence. Accordingly, we propose a novel taxonomy based on the stages in the graph learning pipeline, and highlight the processing methods for different data structures in the graph data, \textit{i.e.,} topology, feature and label.
Furthermore, we analyze some potential problems embedded in graph data and discuss how to solve them in a data-centric manner.
Finally, we provide some promising future directions for data-centric graph learning.
\end{abstract}


\section{Introduction}
\label{sec:introduction}

\IEEEPARstart{R}{ecent} advancements in the non-Euclidean domain draw substantial attention from the artificial intelligence (AI) community.
Graphs, as typical non-Euclidean data, are ubiquitous in the real world and have been widely used in many fields, such as recommendation, security, bioinformatics, etc.
Over the past decade, the progress of graph-related research has been propelled by innovations in graph models, ranging from graph kernels~\cite{Kernel1, Kernel2} to graph embeddings~\cite{Deepwalk, Node2vec}, and culminating in the latest advancements represented by graph neural networks (GNNs)~\cite{GCN, GAT}.
Conversely, little research has been directed toward the intrinsic aspects of graph data, including quality, diversity, security, and so on.

Generally, the revolutions in AI have consistently been initiated by the availability of vast amounts of high-quality data, subsequently followed by powerful models.
A notable example is the success of ImageNet~\cite{Deng2009ImageNetAL}, which significantly contribute to the development of deep convolutional neural networks, \eg, AlexNet~\cite{Krizhevsky2012ImageNetCW} and ResNet~\cite{He2015DeepRL}.
Similarly, the importance of graph data in GNNs learning cannot be overstated. Graphs naturally represent complex relationships and interactions, making them crucial for enhancing the performance of GNNs. For example, the construction of graphs has a substantial impact on the
\IEEEpubidadjcol
performance of graph learning. Even when employing the same GNN model and training methods, the choice between weighted and unweighted graphs, directed and undirected graphs, as well as heterogeneous and homogeneous graphs can result in vastly different final model performance.
As the significance of data becomes increasingly acknowledged, recently, the attention of the AI community has shifted from model-centric approaches to data-centric ones~\cite{data_centric_survey1, data_centric_survey2}.

The emerging data-centric AI emphasizes producing suitable data to improve the performance of a given model.
{\color{blue}}Specifically, data-centric graph learning focuses on optimizing the quality and applicability of graph data, utilizing methods such as graph augmentation~\cite{glcn,TO-GCN,pro-gnn,AdaEdge,GNNGuard,TADropEdge,PTDNet,pmlr-v119-zheng20d,Fairdrop,Eland,GIB, InvRat, GSAT, GREA, DIR, CIGA, GIL, SpCo,SPAN,Ghose2023}, graph sampling~\cite{GraphSAGE,Cluster-GCN,Parallelized_Graph_Sampling, fastgcn,LADIES,graphsaint,ying2018graph,HetGNN}, and feature engineering~\cite{yang2021graph,velivckovic2018deep,feng2019graph,you2020graph,thakoor2021large,wang2020nodeaug,xu2022graph,verma2019manifold}, etc. By properly modifying graph data, these approaches address critical challenges such as sparsity and overfitting, ultimately leading to more effective and efficient graph models. 
``\textit{How to process the graph data to unlock the full potential of graph models?}''
A well-informed answer can help us understand the relationship between graph data and graph models.
However, unlike Euclidean data such as images and tabular data, the irregular nature of graphs poses several questions for data-centric graph learning:
Firstly, \textit{when should we modify graph data to benefit graph models?} Data modification may occur in different stages of graph learning. For example, we can heuristically perturb the edges before training, while we can also estimate new graph structures from the node representations during training.
Secondly, \textit{which part of the graph data should we modify?} Graph data involves various structures, including edges, nodes, features, and labels, each of which plays an important role in graph representation learning.
Thirdly, \textit{how to prevent the graph models from being affected by the problematic graph data?} Graph data may inevitably introduce noise and bias, due to the manually defined relations and features, which makes the models untrustworthy.

\begin{figure*}
\centering
\includegraphics[width=\textwidth]{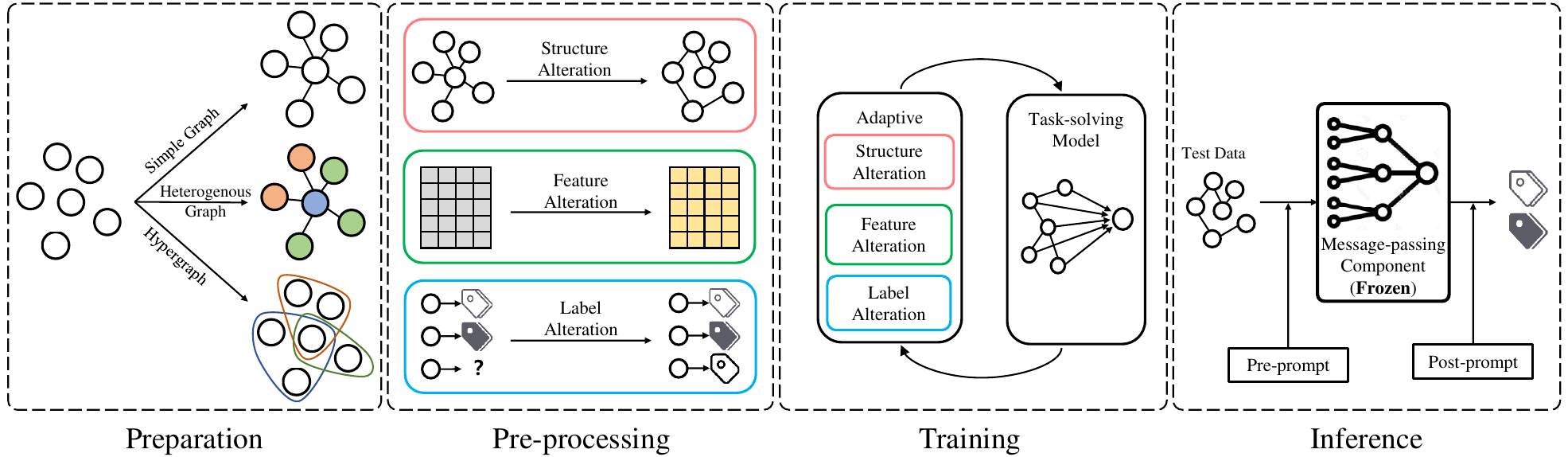}
\caption{Pipeline of data-centric graph learning. The first step is to construct different graphs from the data as needed. The graph structure, node features, or labels are then pre-processed to facilitate the learning of graph models. During the training phase, graph data is collaboratively processed with the task-solving model to improve its performance. Ultimately, prompts are designed to imbue the graph models with enhanced predictive capabilities during the inference stage.
}
\label{fig: framework}
\end{figure*}

This survey systematically reviews and categorizes existing graph learning methods from the data-centric perspective.
In particular, to answer the first question, we divide the graph learning process into four stages: preparation, pre-processing, training, and inference, as illustrated in Figure~\ref{fig: framework}. We discuss the significance of each stage for graph data.
Next, we further categorize existing methods from a structural perspective to address the second question. Specifically, we consider how to handle the topology, features, and labels of graph data, respectively.
Finally, we analyze the potential problems in existing graph data, including vulnerability, unfairness, selection bias, and heterophily. We further discuss how to solve these issues in a data-centric way.

\textbf{Related Surveys.}
Currently, there is some literature on data-centric AI \cite{data_centric_survey1, data_centric_survey2}. However, they mainly focus on the Euclidean domain, and there is little discussion about non-Euclidean data.
Additionally, there are many surveys on the topic of model-centric graph learning. For example, Cui \etal~\cite{NE_survey} summarizes network embedding methods, 
Wu \etal~\cite{GNN_survey} divides GNNs into four representative frameworks,
and Bronstein \etal~\cite{Geometric_survey} studies the equivariant and invariant models on the geometric data.
These surveys introduce various powerful graph models but are orthogonal to our data-centric survey.
\IEEEpubidadjcol
On the other hand, there are also works covering some specific data-centric approaches, \eg, graph augmentation~\cite{augmentation_survey}, graph sampling~\cite{sampling_survey}, and graph structure learning~\cite{structurelearning_survey}, which can be seen as a part of our survey.
The recent work~\cite{DCGL} also provides a review for data-centric graph learning. However, it does not distinguish the different stages of data-centric graph learning.

\textbf{Contributions.} The contribution of this paper is summarized as follows:
\begin{itemize}
    \item \textbf{Novel Taxonomy.} We categorize existing data-centric graph learning methods by the stages in the graph learning pipeline, including pre-processing, training, and inference. For each stage, we introduce its goal and importance for data-centric graph learning.
    \item \textbf{Multiple Perspectives.} We highlight how to process different data structures in the graph data, including topology, feature, and label, to unlock the potential of given graph models.
    \item \textbf{Comprehensive Discussion.} We analyze the potential influence of problematic graph data on the graph models and discuss how to alleviate these problems in a data-centric manner.
    Moreover, we suggest three possible future directions for data-centric graph learning, which may benefit the development of this field.
\end{itemize}

\textbf{Organization.} The rest of this survey is organized as follows:
Section~\ref{sec: preparation} outlines the background of data-centric graph learning, and describes how graph data is manually processed.
Sections~\ref{sec: proprocess}-\ref{sec: inference} introduce the data-centric graph learning methods in the pre-processing, training, and inference stages, respectively.
Section~\ref{sec: problematic} introduces the potential problems of graph data and discusses how to deal with these issues.
Finally, Section~\ref{sec: future} provides a summarization of this paper and poses several promising future directions.

\section{Definition and Background}
\label{sec: preparation}

The first occurrence of the term ``graph'' dates back to 1878, when J. J. Sylvester~\cite{sylvester1878application} used it to establish a connection between mathematics and chemical structures.
Before that, graph theory can be traced back to 1735, when Euler~\cite{euler1741solutio} solved the Seven Bridge of K\"{onigsberg} problem and proved that it is impossible to walk through all seven bridges exactly once.
There are also some other significant graph theory problems~\cite{gross2003handbook}, such as Diagram-Tracing Puzzles, Four-Color Problem, etc.

A graph consists of different nodes and their connections, which can be defined as:
\begin{definition}
    A graph is represented as $G=(V, E)$, where $V$ is the vertex set and $E$ is the edge set.
    Let $|V|$=$N$, the adjacency matrix of graph $G$ is represented as $\mathbf{A} \in \{0,1\}^{N \times N}$, where $A_{ij}=1$ if there is an edge between nodes $i$ and $j$, and $A_{ij} =0$ otherwise.
    A graph including node attributes is defined as an attributed graph, where $\mathbf{X} \in \mathbb{R}^{N \times d}$ is the node feature matrix and the $i$-th row $\mathbf{X}_{i}$ indicates the attributes of node $v_i$. The Laplacian matrix of graph $G$ is denoted as $\mathbf{L}$, and $\hat{\mathbf{L}}=\mathbf{I}_N-\mathbf{D}^{-1/2}\mathbf{AD}^{-1/2}$ is the normalized graph Laplacian matrix with $\mathbf{D}$ the diagonal degree matrix.
\end{definition}

{\color{blue}}

\begin{definition}
Data-centric graph learning aims at improving the quality and utility of graph data (such as topology, features, and labels) across various stages of graph learning (including preparation, pre-processing, training, and inference). Model-centric graph learning, which prioritizes model architecture improvements while assuming the dataset is fixed, often overlooks the complexity, scope, and relevance of the data. In contrast, data-centric approaches emphasize refining the data itself, with the model remaining relatively unchanged. Consequently, data-centric graph learning tends to be more adaptable and transferable across different problems. 
\end{definition}

Building upon the graph definition, we can abstract the real-world objectives and their relations into a graphical representation.
For example, in the K\"{onigsberg} problem, Euler simplifies the islands into nodes and bridges into edges.
In contrast to naturally occurring images and languages,
creating graphs demands significant artificial priors. Consequently, graph construction stands as the \textbf{first step of data-centric graph learning}.

Initially, the definition of edges contains meaningful semantics.
On the one hand, different definitions of edges convey disparate information.
Consider the example of a molecular graph where atoms serve as nodes. 
Edges can be defined as Chemical bonds between atoms, reflecting their chemical properties. Alternatively, edges also can be defined as connections between atoms within a specific radius, \ie, $K$-Nearest Neighbor ($K$NN) graphs, representing the positional information in Euclidean space. 
On the other hand, various auxiliary information introduces diverse edge types as shown in Figure \ref{fig: framework}. For instance, directed graphs consider the directions of edges, heterogeneous graphs~\cite{Heterogeneous_survey} cover different relations between nodes, hypergraphs~\cite{Hypergraph} capture high-order correlations, and temporal graphs~\cite{Temporal_graph} encode the time stamps of edges.

In addition to the definition of edges, there are also various definitions of node features.
For example, within text-attributed graphs, textual information serves as node features.
In the Planetoid datasets~\cite{sen2008collective}, the text is represented as the bag-of-words, while in the OGB datasets~\cite{hu2020open}, the text representations are encoded by the pre-trained language models.
These two different definitions of node features result in a significant performance gap.
Therefore, how to process the node features to improve the performance of graph models is also crucial for data-centric graph learning.

\section{Pre-processing Stage}
\label{sec: proprocess}

\begin{table*}[t]
  \centering
  \caption{Taxonomy and representative works of data-centric graph learning.}
    \begin{tabular}{ccll}
    \toprule
    \multicolumn{1}{l}{Perspective} & Method & Stage & Category \& Reference \\
    \midrule
    \multirow{9}[18]{*}{Topology} & Graph Reduction & $(\S~\ref{sec: graph_reduction})$ Pre-processing & \makecell[l] {Sparsification~\cite{cutsparsifier, benczur2002randomized, fung2011general, srinivasa2020fast,ahmed2020graph,peleg1989graph,chen2022weighted, filtser2021graph,kapralov2020fast,ye2021sparse,ioannidis2020pruned,li2022graph, li2020sgcn, sadhanala2016graph}, Coarsening~\cite{chen2022graph,chen2005matrix, muresan2008analysis, karypis1995multilevel,loukas2018spectrally,jin2020graph,cai2021graph,huang2021scaling}, \\ Condensation~\cite{jin2021graph,zhao2020dataset,jin2022condensing,zheng2023structure,liu2022graph,xu2023kernel,yang2023does,liu2023graph}} \\
\cmidrule{2-4}          & Graph Generation & $(\S~\ref{sec: graph_generation})$ Pre-processing & \makecell[l]{Node~\cite{li2018learning,graphrnn,molecularrnn,graphdf,graphgen}, Adjacency matrix~\cite{digress, EDGE}, Embedding~\cite{vgae, gnf}, \\ Eigenvalue and eigenvector~\cite{spectre, GSDM}} \\
\cmidrule{2-4}          & \multirow{2}[4]{*}{Graph Augmentation} & $(\S~\ref{sec: graph_augmentation})$ Pre-processing & Dropping~\cite{you2020graph,GRACE,GRAND,GPT-GNN,MGAE,DropEdge}, Subgraph~\cite{MoCL,SubMix,MEvolve}, Diffusion~\cite{graph_diffusion,MVGRL,MV-GCN} \\
\cmidrule{3-4}          &       & $(\S~\ref{sec: adaptive_aug})$ Training &  
Structure~\cite{glcn,TO-GCN,pro-gnn,AdaEdge,GNNGuard,TADropEdge,PTDNet,pmlr-v119-zheng20d,Fairdrop,Eland,GIB, InvRat, GSAT, GREA, DIR, CIGA, GIL}, Spectrum~\cite{SpCo,SPAN,Ghose2023}, Automated~\cite{you2021graph,GCA,LG2AR,AutoGDA,GraphAug}\\
\cmidrule{2-4}          & \multirow{2}[4]{*}{Graph Sampling} & $(\S~\ref{sec: graph_sampling})$ Pre-processing & Random~\cite{GraphSAGE,Cluster-GCN,Parallelized_Graph_Sampling}, Importance~\cite{fastgcn,LADIES,graphsaint,ying2018graph,HetGNN} \\
\cmidrule{3-4}          &       & $(\S~\ref{sec: adaptive_sampling})$ Training & Node-wise~\cite{VRGCN,pass,GCN-BS,Thanos,ANS-GT}, Layer-wise~\cite{ASgcn,MVS-GNN}  \\
\cmidrule{2-4}          & \multirow{2}[4]{*}{Graph Curriculum Learning} & $(\S~\ref{sec: graph_curriculum})$ Pre-processing & Node-level~\cite{CLNode,MTGNN,TuneUp,DiGCL,HSAN,DualGCN,GNN-CL,GCN-WSRS}, Graph-level~\cite{CurGraph,CuCo,SMMCL,HACL,CHEST} \\
\cmidrule{3-4}          &       & $(\S~\ref{sec: self_paced})$ Training & Edge~\cite{DSP-GCN, SPC-GNN, li2021robust, chen2023self}, Node~\cite{zhou2023self, zhou2021self, huang2019similarity}, Graph~\cite{gao2018self, li2022mining, zhou2018sparc} \\
\cmidrule{2-4}          & Graph Structure Learning & 
$(\S~\ref{sec: structure_learning})$ Training &
Discrete~\cite{ma2019flexible, zhang2019bayesian, elinas2020variational, pal2020non, vib-gsl, franceschi2019learning, gsebo, kazi2020differentiable}, weighted~\cite{idgl, hgsl, GAT, chen2019graphflow, chen2019reinforcement, on2020cut, yang2018glomo, huang2020location, li2018adaptive, henaff2015deep, liu2019contextualized, liu2020retrieval, jiang2019semi, pro-gnn, yang2019topology, glcn, glnn} \\
    \midrule
    \multirow{5}[8]{*}{Feature} & Feature Augmentation & $(\S~\ref{sec: feature_augmentation})$ Pre-processing & 
    Pre-Attribute ~\cite{yang2021graph,velivckovic2018deep,feng2019graph,you2020graph,thakoor2021large,wang2020nodeaug,xu2022graph,verma2019manifold}, Attribute-Free~\cite{Deepwalk,word2vec,Node2vec,wang2016structural}\\
\cmidrule{2-4}          & Position Encoding & $(\S~\ref{sec: position_encoding})$ Pre-processing & Absolute~\cite{GT, SignNet, SAN,RandomFeature,RWFeature}, Relative~\cite{PGNN, DEGNN, Graphormer, PEG} \\
\cmidrule{2-4}          & Feature Selection & $(\S~\ref{sec: feature_selection})$ Training &  Task-independent ~\cite{graphLasso,reg1,reg2,reg3,fsgnn,wrapper1} , Task-specific ~\cite{fsgnn2,wrapper2} \\ 
\cmidrule{2-4}          & Feature Completion & $(\S~\ref{sec: feature_com})$ Training & Homogeneous~\cite{taguchi2021graph, chen2020learning, jin2022amer, spinelli2020missing}, Heterogeneous~\cite{jin2021heterogeneous, wang2022heterogeneous, he2022analyzing, zhu2023autoac, li2023hetregat} \\
\cmidrule{2-4}          & Graph Prompt & $(\S~\ref{sec: inference})$ Inference & Pre-prompt~\cite{guo2023data, allinone, huang2023prodigy}, Post-prompt~\cite{Liu2023GraphPromptUP, sun2022gppt} \\
    \midrule
    \multirow{3}[6]{*}{Label} & Label Mixup & $(\S~\ref{sec: mix_up})$ Pre-processing & Mixup~\cite{han2022g,airoldi2013stochastic,navarro2023graphmad,park2022graph,wang2021mixup}, Knowledge Distillation~\cite{guo2023boosting, he2022compressing, yang2020distilling, tan2022double, yang2021extract, guo2022alignahead,deng2021graph, yang2023learning, zhang2022multi, wu2022knowledge, dong2023reliant, he2022sgkd} \\
\cmidrule{2-4}          & Pseudo Labeling & $(\S~\ref{sec: pseudo_labeling})$ Training &
% Fine-tune~\cite{zhan2021mutual}, Re-train~\cite{li2018deeper, sun2020multi, li2023informative}
Multi-stage\cite{li2018deeper, zhan2021mutual, sun2020multi, li2023informative}
\\
\cmidrule{2-4}          & Active Learning & $(\S~\ref{sec: active_learning})$ Training &  
Node-independent~\cite{AGE,ANRMAB,ActiveHNE,SmartQuery,Alg},
Node-correlated ~\cite{GPA,DAG,ALLIE,BIGENE,Grain,SAG,IGP,JuryGCN,FeatProp,LSCALE,ScatterSample,Seal,MetAL} \\
    \bottomrule
    \end{tabular}
  \label{tab: taxonomy}
\end{table*}

In this section, we will discuss data-centric methods at the pre-processing stage, which aim to heuristically modify graph data without leveraging the information from task-solving graph models.
Specifically, we consider different data structures in each subsection, including topology, features, and labels.
We also introduce how to add, delete, or change each type of data in detail.

\subsection{Topology}

Topology is the most important part of graph data, representing its structural information.
In this section, we first introduce how to enrich the topology information of graphs, including augmentation and generation. 
Next, we refer to the topology reduction, which removes some redundant edges and nodes in the topology.
Finally, we present curriculum learning and sampling, aiming to speed up the training of graph models by changing the original graph data distribution.

\subsubsection{Graph Augmentation}
\label{sec: graph_augmentation}

Due to the scarcity and sparsity of graph data, it is difficult for graph models to fit the underlying distribution of graph data and easily fall into local optimum.
Therefore, it is important to enrich the topology information in a low-overhead manner.
Under this circumstance, graph augmentation is proposed to alleviate the over-fitting of graph models by perturbing the graph topology without changing its crucial information.
Here we introduce some off-the-training heuristic methods, including masking, substitution, and diffusion.

\textbf{Edge and Node Masking.}
Stochastically masking the edges or nodes in the graphs is a basic but extensively employed technique for graph augmentation.
DropEdge \cite{DropEdge} first proposes randomly masking edges when training GNNs to alleviate the over-smoothing problem. Subsequently, You \etal~\cite{you2020graph} follow a similar scheme that randomly deletes nodes and their associated edges from the graph.
In addition to dealing with the over-fitting and over-smoothing issues, these masking methods have been extensively adopted in graph self-supervised learning (SSL).
On the one hand, graph contrastive learning~\cite{you2020graph, GRACE, GRAND} leverages these incomplete topologies as different views of a graph and learns to capture their consistency.
On the other hand, some generative graph SSL methods~\cite{GPT-GNN, MGAE} treat the masked edges and nodes as self-supervised signals. By reconstructing the masked information, these methods can learn meaningful graph representations.

\textbf{Subgraph Substitution.}
The masking-based methods can only affect the local topology of a graph, overlooking its global structures. In response, subgraph substitution methods have been introduced to replace the specific substructures within graphs as a means of augmentation.
MoCL \cite{MoCL} employs biomedical knowledge to augment molecular graphs by substituting important substructures, such as functional groups, which injects domain-specific insights into graph augmentation.
SubMix \cite{SubMix} utilizes importance sampling to extract and exchange the connected and clustered subgraphs from a pair of graphs, which effectively fuse their topologies.
M-Evolve \cite{MEvolve} uses motifs to augment graph data by selecting target motifs and adjusting edges within them.

\textbf{Graph Diffusion.}
Graph diffusion \cite{graph_diffusion} provides a way to enrich the topology information by aggregating the neighbors at different distances, which can be formulated as:
\begin{equation}
\tilde{\mathbf{A}}=\sum_{k=0}^{\infty}\theta_k\mathbf{T}^k,
\end{equation}
where $\tilde{\mathbf{A}}$ is the augmented adjacency matrix,  $\mathbf{T}=\mathbf{A}\mathbf{D}^{-1}$ is the random walk transition matrix, $k$ is the distance, and $\theta_k$ denotes the weight coefficients.
There are two special examples of graph diffusion: Personalized PageRank (PPR) and heat kernel.
Specifically, PPR uses the random walk transition matrix with coefficients $\theta_k=\alpha(1-\alpha)^k$ and 
heat kernel uses $\theta_k=e^{-s}\frac{s^k}{k!}$ to exponentially reduce the influence of remote nodes, where $\alpha$ and $s$ are hyperparameters.

The aforementioned graph augmentation methods are intuitive, concise, and provide adequate performance in most cases.
Additionally, they can be aggregated to form a complex enhancement strategy, which will be further discussed in Section~\ref{sec: adaptive_aug}.

\subsubsection{Graph Generation}
\label{sec: graph_generation}

Although graph augmentation can initially enrich the topology information, it inevitably introduces noise and hurts the model performance.
As an advanced approach, graph generation~\cite{graph_generation_survey1, graph_generation_survey2} aims to generate graph samples with high quality and diversity.
In this section, we introduce some representative graph generation methods based on the varied generated data, including sequence, embedding, adjacency matrix, and eigenvector.

\textbf{Node Sequence}.
Simplifying a graph into a sequence is the first idea of graph generation, resulting in the autoregressive graph generation methods.
Generally, autoregressive methods aim to generate graphs node-by-node based on the pre-sampled ordering.
However, due to the non-uniqueness and high-dimensional nature of graphs, utilizing node ordering as input needs to consider the permutation-invariance property.
To address this challenge, Li \etal~\cite{li2018learning} randomly sample node orderings from the full permutation of the nodes to guarantee the invariance.
This strategy is effective but inefficient.
Subsequently, GraphRNN~\cite{graphrnn}, MolecularRNN~\cite{molecularrnn}, GraphGen~\cite{graphgen}, and GraphDF \cite{graphdf} employ Breadth-First Search (BFS) or depth-first search (DFS) to guarantee the uniqueness of node orderings and maintain high scalability.

\textbf{Adjacency Matrix.}
In addition to sequence generation, another natural idea is to directly generate the adjacency matrix of the graph.
Methods belonging to this category, \eg, DiGress \cite{digress} and EDGE~\cite{EDGE}, mainly use the emerging diffusion model as the generation framework.
Typically, they first gradually add Gaussian noise into the nodes or edges in the adjacency matrix.
After that, a standard de-noising process is emplyed to recover the adjacency matrix from the noisy graph structures.
A major advantage of this process is that the generation is permutation-equivariant and effective in generating high-quality small graphs.

\textbf{Node Embedding.}
Generating adjacency matrices of graphs is usually time-consuming and cannot scale to large graphs.
One possible solution is to generate the graph in an indirect way. For example, the adjacency matrix can be represented by the node embeddings: $\mathbf{A} = \mathbf{H} \cdot \mathbf{H}^{\top}$.
In this way, we only need to generate a small tensor $\mathbf{H}\in\mathbb{R}^{N \times d}$, rather than the large adjacency matrix $\mathbf{A}\in\mathbb{R}^{N \times N}$, where $d \ll N$.
For example, VGAE \cite{vgae} and GNF \cite{gnf} first utilize GNNs to learn the node representations and then map them into a multivariate Gaussian distribution, where the mean and variance are defined by the node representations.
In the training process, these methods will reconstruct the graph structure via a link prediction loss function.
In the generation process, these methods directly sample node representations from the learned multivariate Gaussian distribution to generate the graph topology.
Although generating node embeddings can scale to large graphs, it still suffers from the high computation cost as it needs to post-process the graphs for alignment.

\textbf{Eigenvalue and Eigenvector.}
Another indirect graph generation method is to generate the eigenvalues and eigenvectors of graphs.
SPECTRE \cite{spectre} and GSDM \cite{GSDM} are the two representative methods, which leverage generative adversarial network and diffusion model to generate the eigenvalues and eigenvectors of graphs, respectively.
Since the spectrum of a graph encodes its global structure, these generation methods can capture the shape information of graphs and therefore result in better generative performance.

\subsubsection{Graph Reduction}
\label{sec: graph_reduction}

The inherent irregular and non-Euclidean structure of graphs prevents the graph models from scaling to large scales~\cite{wu2020comprehensive,zhang2019graph}.
To reduce the computational complexity and improve scalability, one possible way is to reduce the redundant nodes and edges in the graphs without changing the crucial structural information.
Therefore, the graph models trained on the original and reduced graphs will have similar performance.
Graph reduction can be divided into three categories: edge reduction, node reduction, and co-reduction.

\textbf{Edge Reduction.}
Removing the redundant edges in a graph is known as the graph sparsification method, which aims to get a sparsified graph $G_{s} = (V, E_{s})$ by removing some edges of the original graph $G$, where $E_{s} \subset E$.
In general, $G_{s}$ should preserve some crucial properties of $G$, including cut weights, spectral similarity, and shortest path, resulting in three types of graph sparsification methods: cut sparsification~\cite{cutsparsifier, benczur2002randomized, fung2011general}, spectral sparsification~\cite{srinivasa2020fast}, and spanner~\cite{ahmed2020graph}.

Cut sparsification aims to reduce edges while preserving the value of the graph cut $C = (V^{\prime}, V - V^{\prime})$, which divides a graph into two subgraphs with node sets $V^{\prime}$ and $V - V^{\prime}$, respectively.
The value of a graph cut $w(C)$ is defined as the total weight of the edges with the endpoint on each side of a cut.
Generally, $G_{s}$ is a $\epsilon$-cut sparsifier of $G$ if it satisfies the following equation for arbitrary graph cut:
\begin{equation}
    (1 - \epsilon)w_{G}(C) \le w_{S}(C) \le (1 + \epsilon)w_{G}(C),
\end{equation}
where $\epsilon \in (0, 1)$.
This property has been widely used in many graph theories, such as graph connectivity, the maximum flow problem, and the minimum bisection problem.

Spectral sparsification ensures the original and sparse graphs have similar quadratic forms, \ie, smoothness.
$G_{s}$ is a $\epsilon$-spectral sparsifier of $G$ if it satisfies the following equation for any vector $\mathbf{q} \in \mathbb{R}^{N \times 1}$:
\begin{equation}
    (1 - \epsilon)\mathbf{q}^{T} \mathbf{L}_{G} \mathbf{q} \le \mathbf{q}^{T} \mathbf{L}_{S} \mathbf{q} \le (1 + \epsilon)\mathbf{q}^{T} \mathbf{L}_{G} \mathbf{q},
\end{equation}
where $\mathbf{q}^{\top} \mathbf{L} \mathbf{q}$ is the quadratic form.
Spectral sparsification preserves the most important spectral property that represents the global structural information.

Spanner is designed for tasks that focus on the distance of node pairs.
$G_{s}$ is a $t$-spanner of $G$ if their shortest paths $d_{S}$ and $d_{G}$ satisfy the following equation for all node pairs $(v_i, v_j), v_i, v_j \in V$:
\begin{equation}
    d_{G}(v_i, v_j) \le d_{S}(v_i, v_j) \le t \cdot d_{G}(v_i, v_j),
\end{equation}
where $t \geq 1$.
The $t$-spanner can capture the underlying graph structure of a network and perform well in some special graphs, such as rings, meshes, trees, butterflies, and cube-connected cycles.~\cite{peleg1989graph}

There are many methods to implement the aforementioned sparsifiers, including sampling~\cite{srinivasa2020fast, sadhanala2016graph,fung2011general} and dynamic streaming~\cite{chen2022weighted, filtser2021graph,kapralov2020fast}.
Sampling methods aim to assign each edge a possibility. A larger possibility means a higher probability of being removed.
Dynamic streaming methods continuously add or remove some edges in a stream until reach the requirement of the sparsification.
Recent studies have also shown great potential to use graph sparsification to alleviate over-fitting and over-smoothing of GNNs~\cite{ye2021sparse,ioannidis2020pruned}, which may be a future direction of research in graph sparsification.

\textbf{Node Reduction.}
Merging a set of closely connected nodes into a super node can also reduce the size of graphs, which is also known as graph coarsening.
The coarsened graph is defined as $G_{s} = (V_{s}, E_{s})$, where $|V_{s}| = n < N$.
The basic approach of graph coarsening is pairwise aggregation, which merges a pair of nodes into a supernode based on some similarity measures~\cite{chen2005matrix, muresan2008analysis, karypis1995multilevel}. In recent years, spectrum-preserving coarsening methods have received more attention because of their superiority in capturing important structural information.

Loukas \etal~\cite{loukas2018spectrally} propose restricted spectral similarity (RSS) to evaluate whether the coarsened graph can learn the spectrum of the original graph, which is defined as:
\begin{equation}
    (1 - \zeta_{k}) \lambda_{k} \le  \mathbf{u}_{k}^{T} \widetilde{\mathbf{L}} \mathbf{u}_{k} \le (1 + \zeta_{k}) \lambda_{k},
\end{equation}
where $\lambda_{k}$ and $\mathbf{u}_{k}$ indictae the $k$-th eigenvalue and eigenvector of $\mathbf{L}$, respectively. 
$ \widetilde{\mathbf{L}} \in \mathbb{R}^{N \times N}$ is the approximation of $\mathbf{L}$, 
which bridges the gap between the original and coarsened graphs.
Intuitively, $\mathbf{L}_{s}$ preserves the information of $\mathbf{L}$ as it approximates the total variation of the eigenvectors, \ie, smoothness.

Many others follow and continue to work on the graph Laplacian, proposing various methods to measure the distance between the coarse graph and the original graph. For example, Jin \etal~\cite{jin2020graph} proposes spectral distances to capture structural differences between graphs.

\textbf{Co-reduction.}
Simultaneously removing the edges and nodes of a graph is defined as co-reduction. The representative work is graph condensation, which synthesizes a condensed graph $G_{s}$ with fewer nodes and edges. The graph models trained on the original and the condensed graphs can have similar performance.

Jin~\etal ~\cite{jin2021graph} introduce the concept of graph condensation, and provide the first framework of graph condensation.
This framework leverages the gradient matching~\cite{zhao2020dataset} method to align the distributions of original and condensed graphs. Specifically, the optimization process can be formulated as follows: 
\begin{gather}
    \mathbf{\Upsilon}_{\mathcal{S}} = \nabla_{\boldsymbol{\theta}} \mathcal{L}\left(f_{\boldsymbol{\theta}}\left(\mathbf{A}_{s}, \mathbf{X}_{s} \right), \mathbf{Y}_{s} \right), \\
    \mathbf{\Upsilon} = \nabla_{\boldsymbol{\theta}} \mathcal{L}\left(f_{\boldsymbol{\theta}}(\mathbf{A}, \mathbf{X}), \mathbf{Y}\right), \\
    Dis\left(\mathbf{\Upsilon}_{\mathcal{S}}, \mathbf{\Upsilon}\right)=\frac{\mathbf{\Upsilon}_{\mathcal{S}} \cdot \mathbf{\Upsilon}}{\left\|\mathbf{\Upsilon}_{\mathcal{S}}\right\|\left\|\mathbf{\Upsilon}\right\|},
\end{gather}
where $f_{\boldsymbol{\theta}}$ denotes the graph models parameterized with $\boldsymbol{\theta}$, $\mathcal{L}$ is the loss function, and $\mathbf{\Upsilon}_{\mathcal{S}}$, $\mathbf{\Upsilon}$ are the gradients on the condensed and origina graphs, respectively.
By minimizing the distance $Dis$ between gradients, the condensed graph can preserve the structure and feature information of the original graph.

To overcome the computational inefficiency caused by bilevel optimization, DosCond\cite{jin2022condensing} proposes an aggressive approximation of the gradient matching loss, which only matches the gradients of the initial model, avoiding the need for the inner loop. GCDM\cite{liu2022graph} optimizes the synthetic graph by minimizing the distance between the distribution of receptive fields in the original and synthetic graphs. SFGC\cite{zheng2023structure} solely 
optimizes the node features of the synthetic graph, treating the Laplacian matrix constructed by structure information as an identity matrix.
To improve the suboptimal performance caused by the inexact approximation methods aforementioned, KIDD\cite{xu2023kernel} adopts kernel ridge regression for an exact solution and proposes a series of variants to further boost efficiency. SGDD\cite{yang2023does} constructs the structure of the synthetic graph via a structure learning model integrating both feature and auxiliary information.
GCEM\cite{liu2023graph} also tackles the cross-architecture problem from the spectral perspective. Unlike SGDD, GCEM directly optimizes the eigenbasis by aligning the subspaces represented as the outer product of the eigenbasis of the original and synthetic graph respectively, and then constructs the synthetic graph using the learned eigenbasis and the spectrum of the original graph.

\subsubsection{Graph Curriculum Learning}
\label{sec: graph_curriculum}

Curriculum learning~\cite{CL_survey} aims to imitate the human learning process, advocating that the model starts learning from easy samples and gradually advances to complex samples.
Graph Curriculum Learning (CuL)~\cite{curriculum_survey} can benefit the convergence of graph models and improve the generalization ability.
Most graph CuL methods have two important functions: difficulty measurer and training scheduler.
The former evaluates the difficulty of the training data to give the learning prior, and the latter decides how to learn from easy to hard samples.
According to the data structures used in these two functions, we divide existing methods into two categories: node-level and graph-level.

\textbf{Node-level CuL}.
The basic idea is to determine the difficulty of nodes utilizing some statistical information.
For example, CLNode~\cite{CLNode} uses node features, node structure, and label noise to get the difficulty of training samples. MTGNN~\cite{MTGNN} evaluates the difficulty by the length of the predicted steps, while Tuneup~\cite{TuneUp} uses node degree to measure the difficulty.
Data relationship is also a major perspective in designing the difficult measurer. For example, DiGCL~\cite{DiGCL} and HSAN~\cite{HSAN} leverage contrastive loss to judge difficulty, DualGCN~\cite{DualGCN} employs the cross-review strategy, GNN-CL~\cite{GNN-CL} focuses on homophily and smoothness of node and their neighbors, GCN-WSRS~\cite{GCN-WSRS} emphasizes the correlation between samples.

\textbf{Graph-level CuL}.
Different from the above Node-level Cul, Graph-level Cul incorporates graph information into the difficulty measurer and training scheduler. CurGraph~\cite{CurGraph} calculates the difficulty based on the within-class and between-class distributions of the embeddings of the samples., CuCo~\cite{CuCo} determines difficulty based on the similarity of embeddings between negative and positive samples, and SMMCL~\cite{SMMCL} evaluates difficulty with multimodal consistency and multimodal uncertainty.
Some methods, such as HACL~\cite{HACL} and CHEST~\cite{CHEST}, utilize both properties and data relationships to design the difficulty measurer. HACL employs the complexity of the vocabulary in the sample and the consistency between the sentiment expressed by the vocabulary in the sample and the expected sentiment to evaluate the difficulty. CHEST defines the difficulty measurer on multiple pre-training tasks, namely three subgraph context information tasks related to nodes, edges, meta-paths, and a graph contrastive learning task.

\subsubsection{Graph Sampling}
\label{sec: graph_sampling}

Graph sampling methods focus on selecting the nodes that are important for the training of graph models, which can restrict the memory overhead within a fixed budget and speed up the convergence of graph models.
In this part, we discuss the heuristic sampling methods, which can be further divided into two categories: random sampling and importance sampling.

\textbf{Random Sampling}.
Since nodes in graphs have different numbers of neighbors, directly aggregating the information of neighbors will lead to uncontrollable expenses.
The basic way to control the overhead of graph models is to randomly sample the same number of neighbors for each node, making the time and memory overhead controllable during training. Specifically, GraphSAGE~\cite{GraphSAGE} is the first proposed graph sampling algorithm for GNNs. 
For each node, GraphSAGE randomly samples a fixed number of neighbor nodes with equal probability in each convolutional layer of the model for node feature aggregation. Different from GraphSAGE, Cluster-GCN~\cite{Cluster-GCN} performs random sampling in units of subgraphs. It first divides the original graph into multiple clusters by using some graph clustering algorithms and randomly selects a fixed number of clusters, combined into a subgraph for training.
Parallelize Graph Sampling~\cite{Parallelized_Graph_Sampling} is proposed to accurately and efficiently train large-scale graph data, which also randomly samples nodes when generating subgraphs for training.

In summary, random sampling treats the sampled nodes as obeying a uniform distribution. This method overcomes the limitation of neighborhood explosion when aggregating node features and avoids the out-of-memory issue.

\textbf{Importance Sampling}.
Different from random sampling, in order to sample more informative nodes, importance sampling
assigns each node a different sampling probability.
Importance samplings can reduce the variance caused by sampling, leading to a more stable training process and enhanced model performance, which can be classified into node-wise sampling, layer-wise sampling, and subgraph-based sampling based on previous work~\cite{sampling_survey1,sampling_survey2}. 

Specifically, FastGCN~\cite{fastgcn} and LADIES~\cite{LADIES} are typical layer-wise sampling algorithms. 
FastGCN performs node sampling independently at each layer based on a pre-set probability, with the sampling probability calculated by the degree of the node. It believes that nodes with higher degrees contain more information. Differently, LADIES calculates the layer-dependent laplacian matrix for nodes to obtain the sampling probability of each node. 
GraphSAINT~\cite{graphsaint} is a subgraph-based sampling algorithm, which additionally introduces the calculation method of the sampling probability on the edge. These probabilities are also related to the adjacency matrix of the graph and are independent of the training of the model. 
Additionally, some methods are considered as node-wise sampling by us and they are based on random walk such as PinSage~\cite{ying2018graph} and HetGNN~\cite{HetGNN}, which preferentially sample nodes with a high number of visits after the random walk. We also consider the number of visits as the importance of the node.

In Chapter~\ref{sec: adaptive_sampling}, we will discuss adaptive and learnable sampling methods, which perform better sampling of the original graph as the model is trained.

\subsection{Feature}
In this section, we first delve into feature augmentation, dividing our discussion into two distinct categories: pre-Attribute augmentation and attribute-free augmentation. Following this, we explore position encoding methods, which are crucial for capturing structural information in graph networks. This exploration is further categorized into two types: absolute position encoding and relative position encoding. 

\subsubsection{Feature Augmentation}
\label{sec: feature_augmentation}

By creating or modifying node features, feature augmentation introduces additional, relevant information into the dataset, which helps the model generalize better and avoid overfitting.
Depending on the initial availability of node features in a dataset, graph feature augmentation methods can be divided into two main categories: vanilla methods and feature-free methods. 

\textbf{Pre-Attribute Augmentation. }
In graphs equipped with pre-existing features, augmentation methods aim to enhance the utility and diversity of these features through various intuitive adjustments. These methods can be categorized as follows: feature corruption~\cite{yang2021graph,velivckovic2018deep,feng2019graph}, which introduces controlled noise to the features; feature shuffling, which rearranges the features; feature masking~\cite{you2020graph,thakoor2021large}, which selectively hides certain features; feature addition, which introduces new features; feature rewriting~\cite{wang2020nodeaug,xu2022graph}, which alters existing features; feature propagation, which spreads features across the graph; and feature mixing~\cite{verma2019manifold}, which combines features from different nodes. Further details and examples of these methods can be found in a comprehensive survey on the topic~\cite{augmentation_survey}.

\textbf{Attribute-Free Augmentation. }
For nodes without initial features, several methods have been proposed to generate meaningful features. A prominent approach is the utilization of random walks to capture structural information. Perozzi introduces DeepWalk~\etal~\cite{Deepwalk}, which initiates multiple random walks from each node and uses the sequence of nodes in these walks to generate node embeddings utilizing word2vec~\cite{word2vec}. Building on this, node2vec~\cite{Node2vec} extends DeepWalk by incorporating a flexible probability mechanism for guiding the random walks, thereby offering a more nuanced exploration of the graph structure. 
In contrast, an alternative approach known as SDNE~\cite{wang2016structural} employs an encoder-decoder architecture for node feature learning. In this method, each column of the adjacency matrix is treated as the initial node embeddings and serves as input to the encoder. The model computes its loss by assessing the disparity between these initial embeddings and the embeddings generated by the decoder, effectively capturing the underlying structural essence of the graph.

In general, feature augmentation exhibits diversity and flexibility, allowing for customized enhancements tailored to the specific requirements of a given problem.

\subsubsection{Position Encoding}
\label{sec: position_encoding}

It is well-recognized that the expressive power of message-passing neural networks (MPNNs) is bounded by the 1-Weisfeiler-Lehman (WL) test and cannot distinguish the isomorphism graphs~\cite{GIN}.
To break this limitation, a popular way is to augment the node features with some positional information, known as the positional encodings.
In this section, we introduce two types of position encodings: absolute methods and relative methods.

\textbf{Absolute Position Encoding} (APE). The objective of APE is to assign each node a position representation to indicate its unique position in the whole graph.
A popular approach for APE involves utilizing the eigenvectors of the graph Laplacian, which is first introduced by Dwivedi~\etal~\cite{GT} in the graph Transformer architecture and then adopted by~\cite{Graphormer}.

However, the eigenvectors suffer from the sign- and basis-ambiguity, implying that randomly reflecting the signs or rotating the coordinates still satisfies the definition of eigenvalue decomposition~\cite{SignNet}.
To solve these issues, SAN \cite{SAN} proposes to use the Transformer to learn a rotations-equivariant APE. Due to the permutation-invariant property of self-attention, SAN is invariant to the rotations of eigenvectors' coordinates.
Additionally, SignNet \cite{SignNet} proposes to simultaneously take the positive and negative eigenvectors as input, so as to address the sign-flipping problem. And then use invariant and equivariant GNNs to learn positional representations from the eigenvectors.

There are also other APE methods, such as random features~\cite{RandomFeature} and random walk features~\cite{RWFeature}.
However, they suffer from either poor generalization or limited receptive field, which limits their widespread use in graph models.

\textbf{Relative Position Encoding} (RPE).RPE captures the relational information between two nodes by employing the distances between them as position encodings. The existing methods are categorized into 1D-RPE and 2D-RPE based on the involved dimensionality.

Given a target node, 1D-RPE methods aim to leverage the distance between the anchor nodes and the target node as the positional representations.
The pioneering 1D-RPE method, PGNN~\cite{PGNN}, adopts a strategy of random sampling anchor nodes and utilizing the distances to these anchor nodes as position encodings.
Li~\etal~\cite{DEGNN} further proposes distance encoding, which takes the geodesic distances between nodes as the relative positions and avoids the choice of anchor nodes.

2D-RPE methods frequently function as the inductive bias for graph structures, a key component widely employed in the graph Transformer architecture.
For example, Graphormer~\cite{Graphormer} encodes the information of the shortest path between two nodes as the 2D-RPE and adds it to the self-attention matrix to preserve the structural information, which can be formulated as:
\begin{equation}
    A_{i j}=\frac{\left(\mathbf{h}_i \mathbf{W}_Q\right)\left(\mathbf{h}_j \mathbf{W}_K\right)^T}{\sqrt{r}}+b_{\phi\left(v_i, v_j\right)},
\end{equation}
where $\mathbf{h}_{i}$ and $\mathbf{h}_{j}$ are the representations of nodes $v_{i}$ and $v_{j}$, respectively, $r$ is the dimension of representations, and $\phi\left(v_i, v_j\right)$ indicates the shorest path encoding function.

Generally, 1D-RPEs are more user-friendly, whereas 2D-RPEs provide a more comprehensive set of positional information.
The 1D-RPEs can be transformed into 2D-RPEs. For example, PEG~\cite{PEG} uses the Euclidean distance between eigenvectors to reweight the adjacency matrix, which combines the advantages of both APE and RPE.

\subsection{Label}
 In this section, we explore various techniques designed to augment the existing graph label data and mitigate issues related to overfitting and noisy labels. These include label mixup methods, which blend labels from different instances to create new training examples, and knowledge distillation, where a teacher model is employed to generate labels for unlabeled data. Additionally, we discuss alternative approaches that focus on refining or selectively using labels to improve the training process. These methodologies collectively contribute to the robustness and accuracy of GNNs in handling graph-structured data.
\subsubsection{Label Mixup }
\label{sec: mix_up}

Label mixup combines two different instances including their associated labels as a new instance, and employs the mixed instances to train GNN models. This approach effectively increases the diversity of the training data, helping the model to better handle a wider range of inputs, thereby making the learned models more generalized and less overfitted. 
Label mixup methods in graph learning can be broadly divided into graph-level mixup and node-level mixup based on the mixing object.

\textbf{Graph-level Mixup}. The graph-level mixup involves mixing labels of two distinct graphs. 
For instance, G-Mixup~\cite{han2022g} starts by estimating a graphon~\cite{airoldi2013stochastic} using graphs of the same class and then progresses to interpolating graphons of different classes in Euclidean space. This interpolation creates mixed graphons, from which synthetic graphs are generated through sampling. Concurrently, it blends the labels of these classes using a predefined parameter. 
Besides, GraphMAD~\cite{navarro2023graphmad} integrates topology by performing nonlinear graph mixup within a continuous domain characterized by graphons. Moreover, it employs convex clustering to learn data-driven mixup functions, allowing generated samples to exploit relationships among all graphs rather than just pairs of data. Similar to G-Mixup, it combines labels using a pre-defined weight. 
Meanwhile, Graph Transplant~\cite{park2022graph} combines irregular graphs by transplanting subgraphs between instances. This approach leverages node saliency to select meaningful subgraphs effectively and uses adaptive label assignment, demonstrating improved performance across diverse graph domains.

\textbf{Node-level Mixup}. Conversely, the node-level mixup focuses on mixing labels of different nodes, particularly for the node classification task. A notable approach proposed by Wang et al.~\cite{wang2021mixup} utilizes a two-stage Mixup framework. In the first stage, a standard feed-forward process within GNNs is employed to obtain node representations. Following this, in the second stage, Mixup is applied using representations of each node's neighbors derived from the first stage, and the label is mixed by a hyperparameter weight same as the aforementioned works. 

\subsubsection{Knowledge Distillation}
\label{sec: distillation}
Knowledge distillation~\cite{KD_survey1, KD_survey2} enables a light-weight model (\textit{i.e.}, student) to acquire knowledge by regarding the soft predictions made by a high-capacity model (\textit{i.e.}, teacher) as pseudo labels for unlabeled nodes/graphs. 
In this survey, we focus on offline distillation techniques~\cite{guo2023boosting, he2022compressing, yang2020distilling, tan2022double, yang2021extract, guo2022alignahead,deng2021graph, yang2023learning, zhang2022multi, wu2022knowledge, dong2023reliant, he2022sgkd}, which can be seen as data-centric enrichment of labels. Specifically, a static teacher model is utilized to produce labels for unlabeled data, and these freshly generated labels are then harnessed to train the student model.
% Several notable studies in this domain deserve mention. 
For instance, 
Yang et al.~\cite{yang2020distilling} introduce a knowledge distillation approach tailored for GCNs, utilizing a Local Structure Preserving (LSP) module to ensure topology-aware transfer of knowledge from a teacher model to a compact student model. 
BGNN~\cite{guo2023boosting} sequentially transfers knowledge from multiple GNNs into a student GNN, augmented by an adaptive temperature module and a weight boosting module to enhance learning effectiveness. 
Furthermore, CPF~\cite{yang2021extract} extracts GNN knowledge from a pre-trained GNN model and infuses it into a simpler student model, which combines label propagation and feature transformation, to enhance prediction accuracy while preserving interpretability.

\subsubsection{Others}
Beyond the techniques of label mixup and knowledge distillation, which predominantly generate extra labels for unlabeled sets, there exists a body of work that specifically focuses on modifying~\cite{xia2023gnn, li2021unified, zhong2019graph} or omitting noisy labels~\cite{xu2022better} in the labeled set.
For instance, Zhong et al.~\cite{zhong2019graph} employs a GCN to refine noisy predictions by establishing relationships between high-confidence snippets and low-confidence ones, thereby propagating anomaly information to correct erroneous labels. 
Similarly, GNN Cleaner~\cite{xia2023gnn} corrects noisy labels by generating pseudo labels through label propagation, and then adaptively and dynamically adjusting these labels during training. 
In a different vein, Better With Less~\cite{xu2022better} introduces a novel graph selector that identifies the most instructive data points based on predictive uncertainty and inherent properties of graphs.

\section{Training Stage}
\label{sec: training}

In this section, we introduce the graph data modification method in the training phase, where the data modification module and the task-solving model cooperate with each other to improve performance.
Following Section~\ref{sec: proprocess}, we also consider how to operate (\textit{i.e.,} add, delete, or change) different data structures (\textit{i.e.,} topology, feature and label) in each subsection.
The related methods can be seen in Table~\ref{tab: taxonomy}.

\subsection{Topology} 
\label{sec: adaptive_topology}
In this section, we delve into diverse techniques crafted for modifying the graph structure throughout model training. Initially, we present graph adaptive augmentation, striving to seamlessly integrate augmentation procedures during the training phase. Subsequently, we discuss graph adaptive sampling methods capable of adjusting the sampling strategy according to the current model's performance.
Besides, we also present graph structure learning which endeavors to uncover valuable graph structures from data. Finally, we explore self-paced learning which allows the model to measure the difficulty of instances and determine the training progress according to the current model state.

\subsubsection{Graph Adaptive Augmentation} 
\label{sec: adaptive_aug}

The conventional rule-based augmentation methods may not be sufficient when demanding increased robustness and improved performance, primarily due to their independence from the task-solving model's training process.
Conversely, graph adaptive augmentation techniques integrate augmentation procedures seamlessly during the training phase, which we classify into four distinct categories. Edge-based methods revamp the adjacency matrix guided by loss functions, subgraph-based methods concentrate on extracting a more informative subgraph, spectrum-based methods propose to generate an augmentation from the graph spectrum view, and automated augmentation frameworks introduce to learn optimal augmentation strategy for varied scenarios.

\textbf{Edge-based.}
To ensure that the edge augmentation process is guided by a differentiable loss function, some studies (e.g., GLCN \cite{glcn}, TO-GCN \cite{TO-GCN}, Pro-GNN \cite{pro-gnn}) treat the graph as continuous rather than having exact edges. These works formulate loss functions by incorporating specific constraints, such as smoothness and sparsity, to generate gradients for refining the graph structure.
As an example, Pro-GNN \cite{pro-gnn} proposes a loss function:
\begin{equation}
   \mathcal{L}=||\tilde{\mathbf{A}}-\mathbf{A}||^2_F + \eta ||\tilde{\mathbf{A}}||_1 + \beta ||\tilde{\mathbf{A}}||_* + \rho (\mathbf{X}^T{\hat{\mathbf{L}}}\mathbf{X})+ \gamma \mathcal{L}_{GNN}, 
\end{equation}

where $\tilde{\mathbf{A}}$ is the augmented graph adjacency matrix, and $\hat{\mathbf{L}}$ is the normalized laplacian matrix. Specifically, the term $||\tilde{\mathbf{A}}-\mathbf{A}||^2_F$($||\cdot||_F$ means Frobenius norm) aims to make $\tilde{\mathbf{A}}$ close to $A$. The term $\eta ||\tilde{\mathbf{A}}||_1$ ($||\cdot||_1$ means $L_1$ norm) and $\beta ||\tilde{\mathbf{A}}||_*$ ($||\cdot||_*$ means nuclear norm) ensure sparsity and low-rank properties, respectively. The term $\rho (\mathbf{X}^T\hat{\mathbf{L}}\mathbf{X})$ controls feature smoothness, and $\gamma \mathcal{L}_{GNN}$ is the empirical loss.

Other studies do not directly implement alterations to the graph through gradients; instead, they iteratively generate or remove discrete edges during training. A representative work is AdaEdge \cite{AdaEdge}, where edges are added or removed based on the classification results of their respective nodes. Similarly, GNNGuard \cite{GNNGuard} selectively removes edges between dissimilar nodes that may be malicious. Besides, TADropEdge \cite{TADropEdge} drops edges according to their weights, calculated from the graph spectrum. PTDNet \cite{PTDNet} and NeuralSparse \cite{pmlr-v119-zheng20d} leverage sparsity and low-rank properties to remove task-irrelevant edges. Furthermore, there are also some related works proposed to fulfill different demands, such as fairness (\cite{Fairdrop}), and anomaly detection (\cite{Eland}).

\textbf{Subgraph-based.}
Subgraph-based adaptive augmentation aims to find the most representative and informative subgraph (also known as the graph rationale), like the functional groups in a molecule, to enhance the model in terms of performance, interpretability, robustness, and so on (\cite{GIB}, \cite{InvRat}, \cite{GSAT}, \cite{GREA}). For instance, GIB~\cite{GIB} extracts information from graph structure and node features, adopting an Information Bottleneck (IB) perspective, and encourages the learned representation to contain the minimum amount of information appropriate for downstream prediction tasks. 
Consequently, models trained using this approach demonstrate reduced susceptibility to overfitting and increased robustness against adversarial attacks.
Moreover, GSAT~\cite{GSAT} proposes stochastic attention to reduce information from the input graph, thus achieving better generalization and interpretation. GREA~\cite{GREA} introduces environment replacement augmentation, which identifies the rationale and separates it from the graph, then substitutes the remaining part, which they call the environment, to generate augmented data. Besides, several related methods garner attention within the out-of-distribution (OOD) research field (\cite{DIR}, \cite{CIGA}, \cite{GIL}). These approaches are devised to separate and leverage causal and non-causal, or invariant and variant subgraphs, with the aim of enhancing OOD generalization.

\textbf{Spectrum-based.}
It is preferred to consider global information to better preserve graph properties when generating an augmentation~\cite{graph_diffusion}. Graph spectrum is such a tool that naturally incorporates global graph properties including clusterability, connectivity, d-regularity, etc., offering an effective way to gain rich information with respect to the overall structure of the graph.
Specifically, the graph spectrum $\mathbf{\Lambda}$ is calculated by an eigendecomposition procedure:
\begin{equation}\hat{\mathbf{L}}=\mathbf{U\Lambda U^T},
\end{equation}
With such a powerful tool, several spectral augmentation methods are proposed. 
SpCo \cite{SpCo} seeks to identify optimal contrastive pairs by amplifying high-frequency amplitudes, thereby introducing distinctions between the original and augmented graphs, while retaining low-frequency amplitudes to preserve invariant information within the graph. Simultaneously, it addresses the optimization problem by treating it as a matrix perturbation through Sinkhorn's Iteration.
SPAN \cite{SPAN} also aims to maximize the spectral difference between two graphs, but the optimization is guided by gradients. Furthermore, SFA \cite{Ghose2023} presents two augmentation methods: spectral graph cropping, which involves the removal of the two smallest non-zero eigenvalues, and graph frequency component reordering, which permutes the eigenvectors associated with the top $k$ eigenvalues.

\textbf{Automated Augmentation.}
The methods mentioned above utilize fixed augmentation strategies. Nonetheless, the same augmentation strategy may not be suitable for diverse datasets. Consequently, a significant portion of research is transitioning towards automatic augmentation, which entails the learning of the optimal augmentation strategy during the model training process. For example, JOAO \cite{you2021graph} sets up a bi-level optimization procedure to train the encoder and augmentation strategy simultaneously.  
Besides, many methods take advantage of reinforcement learning (RL) in data augmentation. LG2AR \cite{LG2AR} learns a policy to assign probabilities to a set of strategies (edge dropping, feature masking, etc.) and samples from them in each training epoch.   
AutoGDA \cite{AutoGDA} introduces a community-customized graph data augmentation method that employs an RL-agent to determine the optimal strategy for each community within a given graph.
GraphAug \cite{GraphAug} focuses on providing label-invariance augmentations for different graphs, implemented by an RL-agent which takes graph embedding as input and samples a strategy.

\subsubsection{Graph Adaptive Sampling}% 
\label{sec: adaptive_sampling}

In this subsection, we discuss graph adaptive sampling algorithms, which can adjust the sampling strategy based on the current model's performance, facilitating more effective utilization of graph information during the training process and thus improving the model's performance.
According to the strategy and scope during node sampling, we currently classify these methods into node-wise sampling and layer-wise sampling as mentioned in ~\ref{sec: graph_sampling}.

\textbf{Node-wise Sampling.} Node-wise sampling is a fundamental sampling technique that samples a fixed number of neighboring nodes for each target node layer by layer.

In particular, VR-GCN~\cite{VRGCN} performs a fixed number of random sampling of neighbor nodes like GraphSAGE. However, we classify it as a graph-adaptive sampling method because it retains and utilizes intermediate node information that evolves as the model learns to approximate unsampled nodes. This approach enables the model to gather a relatively greater amount of node information, ultimately enhancing model performance with a small number of samples.
PASS~\cite{pass} directly utilizes gradient information and task performance loss to train a sampling policy. Since the sampling operation is non-differentiable, PASS proposes to learn from the gradients to determine which neighbors provide valuable information and are therefore assigned high sampling probabilities.
Besides, GCN-BS~\cite{GCN-BS} and Thanos~\cite{Thanos} consider the sampling problem from another perspective.  
GCN-BS reformulates the optimization of sampling variance by treating it as a bandit problem, establishing a connection between sampling variance and both the reward and the regret. To achieve adaptive sampling probabilities, GCN-BS calculates the reward and subsequently updates the sampler following back propagation.
Based on GCN-BS, Thanos introduces a novel biased reward function and corresponding regret, while also loosening the model's data-related assumptions. Experimental results demonstrate that Thanos surpasses GCN-BS in terms of approximation error and converges at a nearly optimal rate.
Furthermore, ANS-GT~\cite{ANS-GT} is applied to the graph transformer, which develops a framework that combines multiple node-wise heuristic sampling strategies and designs learnable weights for each heuristic strategy to adaptively select the best-performing sampling algorithm during training, making the model achieve the best performance. 

\textbf{Layer-wise Sampling.} Different from the above method, 
layer-wise sampling abandons sampling from nodes and, instead, within the realm of graph convolutional layers, selects a consistent number of nodes at each layer. 
For example, AS-GCN~\cite{ASgcn} is a typical adaptive layer-wise sampling method, which samples nodes for the lower layers based on the nodes in the top layer and enables the sharing of sampled neighbor nodes among different parent nodes. This approach facilitates control over the number of nodes at each layer, preventing excessive expansion. More importantly, the newly proposed layer-wise sampling method in AS-GCN is adaptive and explicitly reduces sampling variance, thereby enhancing the training effectiveness of this approach.
Additionally, MVS-GNN~\cite{MVS-GNN} theoretically analyzes the components of sampling variance and introduces gradient-based minimization of variance sampling. It samples from the optimal importance sampling distribution by computing the gradient norm, ensuring that the sampled nodes consistently maintain the smallest sampling variance.

\subsubsection{Graph Structure Learning} 
\label{sec: structure_learning}

The effectiveness of GNNs in capturing expressive representations is significantly influenced by the quality of the underlying graph-structured data.
However, real-world graph structures are frequently noisy or incomplete. Therefore, Graph Structure Learning (GSL) endeavors to uncover valuable graph structures from data to enhance the learning of graph representations. 
Depending on whether weight information on the edges is considered, existing GSL methodologies can be broadly classified into two categories: learning discrete graph structures and learning weighted graph structures.

\textbf{Learning Discrete Graph Structures.} The methods in this category consider graph structures as random variables, where discrete graph structures can be sampled from a certain probabilistic adjacency matrix. Various techniques, such as variational inference~\cite{ma2019flexible, zhang2019bayesian, elinas2020variational, pal2020non, vib-gsl}, bi-level optimization~\cite{franceschi2019learning, gsebo}, and reinforcement learning~\cite{kazi2020differentiable}, are leveraged to jointly optimize both the probabilistic adjacency matrix and GNN parameters. For example, $\text{G}^{3}\text{NN}$~\cite{ma2019flexible} treats node features, graph structure, and node labels as random variables, then utilizes a flexible generative model to capture their joint distribution for the graph generation. VGCNs~\cite{zhang2019bayesian} aims to maximize the posterior over binary adjacency matrix given the observed data, namely node features and observed node labels. In contrast to variational inference approaches, LDS~\cite{franceschi2019learning} and GSEBO~\cite{gsebo} view the task, i.e., optimizing the graph structure and GNN parameters, as a bi-level optimization problem. Furthermore, DGM~\cite{kazi2020differentiable} employs reinforcement learning to tackle the non-differentiability challenge arising from the edge sampling operation during optimization.

\textbf{Learning Weighted Graph Structures.} Since learning discrete graph structures tends to optimize the adjacency matrix directly based on certain prior constraints on the graph properties, many of these approaches are unsuitable for the inductive learning setting where unseen nodes emerge during the inference phase~\cite{wu2022graph}. Consequently, inspired by attention techniques~\cite{GAT}, a class of methods focuses on learning weighted graph structures, i.e., edge weights between nodes. Based on various similarity measures, such as cosine-based similarity~\cite{idgl, hgsl}, attention mechanisms~\cite{GAT, chen2019graphflow, chen2019reinforcement, on2020cut, yang2018glomo, huang2020location}, kernel-based similarity~\cite{li2018adaptive, henaff2015deep}, many methods take node embedding to learn pairwise node similarity matrices. Some methods further integrate intrinsic edge embeddings~\cite{liu2019contextualized, liu2020retrieval} or edge connection information~\cite{jiang2019semi} into the process of similarity learning. Additionally, graph regularization techniques~\cite{pro-gnn, idgl, yang2019topology, glcn, glnn} directly optimize the graph structure by considering various graph properties, such as smoothness, connectivity, low rank and sparsity. For instance, Pro-GNN~\cite{pro-gnn} learns a refined graph structure from a perturbed graph by leveraging the sparsity, low rank, and smoothness properties of the graph.  
When a new structure is learned through the above methods, completely discarding the original graph structure may result in the loss of valuable information. In light of this, recent work~\cite{li2018adaptive, idgl} suggests utilizing both the learned graph structure and the original graph structure through a linear combination.

\subsubsection{Graph Self-paced Learning}
\label{sec: self_paced}
As a special curriculum learning algorithm, self-paced learning allows the task-solving model to measure the difficulty of instances and determine the training progress according to the current model state~\cite{SPL}. 
Similar to Section~\ref{sec: graph_curriculum}, we categorize existing methods into node-level, link-level, and graph-level methods.

\textbf{Node-level Self-paced Learning.} The basic idea is to determine the training nodes according to the current training state. DSP-GCN~\cite{DSP-GCN} and SPC-GNN~\cite{SPC-GNN} gradually incorporate unlabeled nodes with higher conﬁdence predictions into the training set, while SS-GSELM~\cite{li2021robust} and SPGCN~\cite{chen2023self} according to the loss value of the labeled nodes in each training, nodes with smaller loss values are prioritized for training.			

\textbf{Edge-level Self-paced Learning.} Different from node-level, edge-level self-paced learning gradually introduces the relationships between nodes in the training process. For instance, SCCABG~\cite{zhou2023self} and SCCBG~\cite{zhou2021self} determine the reliability of each edge by an adaptive clustering similarity measure, and then the edges gradually are included in order of reliability. SANE~\cite{huang2019similarity} gradually introduces the semantic relationship between nodes into the network representation learning by considering node similarity.

\textbf{Graph-level Self-paced Learning.} The core insight of such approaches is to gradually determine the context of the center node. For example, SeedNE~\cite{gao2018self}, relying on the sampling probability of nodes, progressively selects difficult negative context nodes to learn better node representations. In contrast, SPGCL~\cite{li2022mining} prioritizes the nodes with the largest mutual information as neighbors. Meanwhile, SPARC~\cite{zhou2018sparc} selects graph contexts for model training based on the number of labels in different classes, gradually focusing on contexts associated with rare classes.

\subsection{Feature}
In this section, we explore the manipulation of graph features during model training, organizing our discussion into two distinct categories: feature completion and feature selection. Feature completion is designed to generate missing node features, addressing incomplete features in graph data. Conversely, feature selection aims to pinpoint highly valuable features during model training.

\subsubsection{Feature Completion} 
\label{sec: feature_com}
The majority of GNNs assume that the node features on the graph are complete, but this assumption is often broken in practical applications, the reasons mainly come from the following aspects~\cite{taguchi2021graph}: (1) machine or human errors during the data collection process; (2) collecting the dataset completely is very costly in practice; (3) many users are not willing to provide complete personal information due to privacy protection. Therefore, to address the problem of incomplete features in graph data, feature completion as a key solution aims to fill in the missing node features in the graph. According to different types of graph data, the existing methods can be roughly categorized into homogeneous and heterogeneous graph-based feature completion.

\textbf{Homogeneous Graph-based Feature Completion.} For homogeneous graphs, $\text{GCN}_{mf}$~\cite{taguchi2021graph} employs the Gaussian mixture model to represent the missing node features. Based on a shared-latent space assumption on graphs, SAT~\cite{chen2020learning} designs a distribution matching GNN architecture for feature completion. Amer~\cite{jin2022amer} develops a novel generative adversarial network to generate missing attributes. GINN~\cite{spinelli2020missing} reconstructs a complete graph by using a GNN-based autoencoder network.

\textbf{Heterogeneous Graph-based Feature Completion.} For more complex heterogeneous graphs, a series of node feature completion methods capable of handling different types of nodes and edges have been proposed ~\cite{jin2021heterogeneous, wang2022heterogeneous, he2022analyzing, zhu2023autoac, li2023hetregat}. For example, AC-HEN~\cite{jin2021heterogeneous} utilizes feature aggregation and structural aggregation to obtain multi-view embeddings for completing missing attributes. HGNN-AC~\cite{wang2022heterogeneous} pre-learns topological embeddings and then utilizes them to guide the feature completion. HGCA~\cite{he2022analyzing} designs an augmentation network that captures the semantic relationships between nodes and attributes to achieve fine-grained attribute completion. AutoAC~\cite{zhu2023autoac} models the attribute completion problem as an automatic search problem for each missing attribute node.

\subsubsection{Feature Selection}
\label{sec: feature_selection}

The cost of model training significantly rises when data used in machine learning algorithms displays sparse features in a high-dimensional space, which we call the curse of dimensionality. As a result, Feature Selection (FS) emerges as one approach to mitigate this challenge.
The goal of FS is to identify highly correlated features with the labels and prioritize them during model training. FS not only helps reduce the computational costs associated with high-dimensional data but also improves model performance by fitting meaningful features.
In graph learning, the frequently utilized FS methods, incorporated into the model training phase, can be classified into two types based on their relationship with downstream tasks: task-independent FS and task-specific FS. 

\textbf{Task-independent FS}. Such methods concentrate on generating superior features and seamlessly integrate with any GNN model or downstream tasks. To start, various works~\cite{graphLasso,reg1,reg2,reg3} center around introducing a regularization objective for feature selection. For example, Graph Lasso~\cite{graphLasso} incorporates a graph regularizer based on Lasso~\cite{tibshirani1996regression} for the feature graph to account for structural information. Further, AsGNNS~\cite{reg2} combines L2, 1/L1-norm regularized attribute selection and GNNs together to extract meaningful features and eliminate noisy ones. 
Except for regularization-based FS, \cite{fsgnn} proposes a task-independent method, which first extends the feature selection algorithm presented via Gumbel Softmax to GNNs to extract features, and then implements a mechanism to rank the extracted features. ADAPT~\cite{wrapper1} introduces a framework for feature selection with the goal of identifying informative features that accurately describe the adaptive neighborhood structure of a network.

\textbf{Task-specific FS}.
In contrast, the task-specific FS methods take the GNN model's task into consideration. Noticing that selective aggregation outperforms default aggregation in node classification tasks, Dual-Net GNN~\cite{fsgnn2} suggests a classifier model trained on a subset of input node features to predict node labels and a selector model that learns to provide the optimal input subset to the classifier for achieving best performance.
Lin \etal~\cite{wrapper2} introduce FS-GCN, an FS method that integrates an indicator matrix into the propagation process of GCN. The optimization of the indicator matrix involves minimizing the cross-entropy loss derived from the semi-supervised node classification task, coupled with a sparsity-based regularization.

\subsection{Label}
In this section, we explore the augmentation of labels during model training, organizing our discussion into two distinct categories: active learning and pseudo labeling. 
Active learning is devised to strategically select the most impactful data and label it for enhancing the task-solving model. 
Additionally, pseudo labeling aims to expand the label set by employing a trained model and assigning pseudo-labels.
\subsubsection{Active Learning}
\label{sec: active_learning}

In real-world scenarios, demanding a large quantity of labeled data to achieve an excellent model is both expensive and unrealistic. Therefore, the concept of Active Learning (AL) has been introduced.
It aims to select the most effective data for the task-solving model from the data set and label it to get the best model performance when the labeling cost is limited. 
While numerous AL methods have been employed in graph data, the majority remains centered on node classification problems. Categorically, the existing AL methods applied to graph data can be segregated into two groups: node-independent and node-correlated AL methods.

\textbf{Node-independent AL}. These methods usually select nodes in unlabeled data for labeling based on metrics and rules. They regard each node selection as an independent process and believe that the current node selection process will not have an impact on other nodes. For example, AGE\cite{AGE} uses three indicators: information entropy, information density, and graph centrality to select the most informative nodes for labeling. 
ANRMAB\cite{ANRMAB} focuses on improving AGE by employing a multi-armed bandit mechanism to dynamically learn weights for balancing the aforementioned three metrics. 
Similar to ANRMAB, ActiveHNE\cite{ActiveHNE} is generalized to heterogeneous graphs.
Besides, SmartQuery\cite{SmartQuery} introduces degree and PageRank informativeness measurements to select nodes. Furthermore, ALG\cite{Alg} introduces a novel metric called Effective Reception Field (ERF), which combines receptive field with node effectiveness measurements, leading to the involvement of more nodes in training when ERF is maximized.

Overall, these methods use greedy ideas to independently select nodes to label based on different measurements, and cannot guarantee that the model achieves long-term optimal performance.

\textbf{Node-correlated AL}. Differing from node-independent AL, such AL methods assume nodes are correlated and consider interactions between nodes.
This idea avoids that the selected nodes are similar and clustered together, which leads to no duplication of information.
Consider a straightforward example where, in node-independent methods, two similar and informative nodes might be chosen and labeled sequentially, resulting in a failure to maximize information acquisition for enhancing model performance within limited labeling costs. In contrast, node-correlated methods address this limitation. Currently, these methods can be broadly categorized into three groups: reinforcement learning (RL)-based, influence function-based, and clustering-based.

Many RL-based methods model active learning on graphs as a Markov decision process. The selection of nodes for marking is regarded as an action, and the performance of the model based on the selected nodes is viewed as a reward, so as to consider the interaction between nodes and obtain the long-term performance of the model. 
Specifically, GPA\cite{GPA} parameterizes the policy network as a GNN and utilizes reinforcement learning to train the policy network.
Subsequently, the network generates a probability for each unlabeled node and then samples a node for labeling based on the probability. Inspired by GPA, DAG\cite{DAG} learns a graph-specific policy based on a universal policy for each graph when solving the transferable problem. The knowledge learned by graph-specific policies is dynamically distilled into the universal policy by minimizing the KL divergence between graph-specific policies.
Similarly, ALLIE\cite{ALLIE} applies an imbalance-aware reinforcement learning based graph policy network to find unlabeled nodes that maximize model performance, and uses a graph coarsening strategy to make it applicable to the large-scale graph. BIGENE\cite{BIGENE} introduces multi-agent reinforcement learning into batch active learning to improve sampling efficiency and considers both node informativeness and diversity.

There are still some methods based on the influence function that consider the information of each node. After message passing, the most effective nodes are selected through the aggregation of node information from a full-graph perspective to obtain long-term effects. For example, Grain\cite{Grain} connects active learning on graphs with social influence maximization and proposes a diversified influence maximization strategy to select nodes. SAG\cite{SAG} introduces the L1-norm of the expected Jacobian matrix as the influence of nodes after k-layer GCN propagation. IGP\cite{IGP} proposes relaxed query and soft label conditions, and selects nodes by maximizing a new criterion called information gain propagation. 
Moreover, the influence function is employed in JuryGCN\cite{JuryGCN} to quantify jackknife uncertainty for each node, and nodes exhibiting high jackknife uncertainty are then selected for active learning.

Clustering-based active learning methods for graph data ensure the diversity of nodes to the greatest extent and avoid redundant information. For example,
FeatProp\cite{FeatProp} uses propagated node features for clustering and labels the nodes at the center of each cluster. Unlike FeatProp, LSCALE\cite{LSCALE} clusters node embeddings in a latent space that contains two key attributes: low label requirements and informative distances. In ScatterSample\cite{ScatterSample}, the uncertainty of all nodes is first calculated, and then the top uncertain nodes are clustered to ensure the diversity of sampling.

Some other different advanced technologies have also been proposed. SEAL\cite{Seal} introduces adversarial learning into graph active learning for the first time. And MetAL\cite{MetAL} evaluates the importance of nodes based on meta-gradients rather than heuristic rules.

In general, no matter what strategy is used (RL-based, influence function-based and clustering-based, etc.), the node-correlated AL methods always do not treat the selection of each node as an independent process. They consider both the informativeness and diversity of nodes to different extents to avoid redundant information and achieve long-term effects.

\subsubsection{Pseudo Labeling}
\label{sec: pseudo_labeling}
With a multi-stage training paradigm, Pseudo Labeling~\cite{li2018deeper, zhan2021mutual, sun2020multi, li2023informative} utilizes a trained model to expand the label set by assigning a pseudo-label, then fine-tunes the trained model or re-trains a new model. For example, Li et al.~\cite{li2018deeper} first present a pseudo-labeled GCN, where the top K high-confidence unlabeled nodes are selected to expand the training set for model retraining. MT-GCN~\cite{zhan2021mutual} takes the pseudo label to accomplish the mutual teaching process across the two GCNs. M3S~\cite{sun2020multi} employs a deep clustering model to assign pseudo-labels. InfoGNN~\cite{li2023informative} considers both informativeness and prediction confidence of pseudo-labeled nodes, to further solve the problem of information redundancy and noisy pseudo-labels in existing methods.

\section{Inference Stage}
\label{sec: inference}
The inference stage refers to the phase where a pretrained graph model is used for the prediction of downstream tasks. Moreover, inference data refers to the graph data utilized during the inference phase of pretrained models. In this stage, we can adjust the inference data to fulfill different trustworthy requirements or reformulate the downstream tasks to align with the pretext tasks ensuring high-quality knowledge transfer. From a data-centric perspective, using prompts to modify the inference data can help obtain the desired objectives without changing model parameters. In this section, we discuss prompt learning methods which are gradually gaining popularity in the realm of graphs. To elaborate, we classify existing graph prompting methods into two categories: pre-prompt and post-prompt, based on whether the task-specific prompts operate before or after the message passing module as shown in Figure\ref{fig: framework}.

\subsection{Pre-prompt}
\label{sec: pre_prompt}

In the case of pre-prompt methods, existing works such as \cite{Jin2022EmpoweringGR, allinone, guo2023data, fang2022universal, zhu2023sgl, huang2023prodigy} modify the input graph data either in terms of topology or node features before message passing to facilitate downstream task adaptation, or they construct a prompt graph to promote the model's adaptation to downstream tasks.

A classical example that applies prompt learning to graph neural networks is AAGOD\cite{guo2023data}, which utilizes prompt learning to modify the graph topology thus achieving adaptation for out-of-distribution (OOD) detection tasks without changing the parameters of GNN backbones. Using a parameterized matrix as a learnable instance-specific prompt, AAGOD superimposes this prompt on the adjacency matrix of the original input graph and reuses
the well-trained GNN to encode the modified graph into
vector representations. The instance-specific prompt helps highlight potential patterns in the in-distribution (ID) graphs, thereby increasing the difference between OOD and ID graphs. 

In a similar vein, a notable contribution in this area is a multi-task prompt method \cite{allinone} which uses prompts to modify the input feature. It bridges the gap between graph pre-training and downstream tasks under the multi-task background. This work initially constructs induced graphs for nodes and edges through $\tau$-hop neighbors, thereby reformulating node-level and edge-level tasks into graph-level tasks. Additionally, it designs prompt tokens for the input graph and modifies the features of each node by weighting all the prompt tokens before message passing. Finally, using the meta-learning paradigm, the prompt parameters are updated for multi-task scenarios.

Another approach to designing prompts for graphs involves utilizing a prompt graph to assist model training. Prodigy \cite{huang2023prodigy} proposes a graph prompt learning design that not only reformulates downstream tasks into a unified template but also introduces a prompt graph with regard to few-shot tuning. In the process of few-shot prompting for downstream tasks, Prodigy first retrieves the corresponding context from the source graph data. Subsequently, it abstracts the corresponding task into a task graph and combines it with the data graph to form a prompt graph. Once the prompt graph is obtained, leveraging a proficiently trained GNN via specifically crafted pre-training tasks enables the acquisition of predicted labels for each data point.

\subsection{Post-prompt}
\label{sec: post_prompt}

As for post-prompt methods \cite{Liu2023GraphPromptUP,sun2022gppt}, task-specific prompts often operate on representations that have already undergone message passing to enable downstream task adaptation.

GraphPrompt \cite{Liu2023GraphPromptUP} is also among the early attempts at prompt learning in the field of graphs. It similarly aims to bridge the gap between pretext tasks and downstream tasks but in a post-prompt manner. This framework begins with pretraining on unlabeled graphs employing a self-supervised link prediction task. It unifies node classification tasks and graph classification tasks into a link prediction form by adding pseudo nodes, thereby eliminating the gap between pretext tasks and downstream tasks. Subsequently, it utilizes learnable prompts to guide each downstream task. This learnable prompt can be understood as a weighted mask to the readout representation. As a result, it can be applied to various tasks, each with a distinct emphasis on different feature channels.

Another pioneering work is GPPT \cite{sun2022gppt}. Unlike GraphPrompt which achieves a form of multi-task unification for tasks at graph-level, node-level, and edge-level, GPPT primarily focuses on node classification tasks. Similarly, GPPT utilizes link prediction as a pretraining task, although it differs in prompt design by concatenating task-specific prompts with node representations to guide adaptation.

\section{Problematic Graph Data}
\label{sec: problematic}

Manually defined and processed graph data inevitably introduces noise and problems.
The aforementioned methods are usually used in general graph data, without considering the specific issues hidden in the graph structure.
In this section, we list a series of commonly introduced problematic graph data and discuss how to deal with these issues in a data-centric approach.

\textbf{Vulnerability.}
Recent advances in graph adversarial learning show that graph structures are vulnerable~\cite{sun2022adversarial}, where a small perturbation in the structures, features, or labels can significantly affect the predictions of the graph models~\cite{zugner2020adversarial}.
However, existing methods focus on designing graph defense models to handle adversarial edges and lack generalization.
The emerging certificate method~\cite{certificate} provides a data-centric way to consistently improve the robustness of data against perturbations, which has been widely used in graph data.
For example, Bojchevski~\etal~\cite{certificate_graph} first propose the verifying certifiable robustness of graph data, improving the robustness by constraining the local and global certificates.
Tao~\etal~\cite{Immunization} further present the immunization method for graph data, which injects protective edges to improve the robustness of graphs against perturbations.

\textbf{Unfairness.}
Existing literature shows that graph models may have inherent prejudice if the training data contains sensitive attributes or specific structures.
Feature fairness requires graph models to make predictions without using sensitive attributes, such as gender and race.
Agarwal~\etal~\cite{fair_augmentation} propose fairness-aware graph augmentation, which utilizes the counterfactual perturbation to make graph models learn invariant representations against sensitive attributes.
Structural fairness refers to the phenomenon that graph models have different accuracy on nodes with different structures, \eg, high-degree and low-degree nodes~\cite{Liu2023ASO}.
GRADE~\cite{GRADE} first finds that graph contrastive learning has better structural fairness than semi-supervised GNNs. Based on this discovery, they propose interpolation-based and purification-based graph augmentations for low- and high-degree nodes.

\textbf{Selection Bias.}
Due to the influence of human factors, the collection of graph data will inevitably introduce selection bias, making the distribution of training data and test data inconsistent.
For example, existing molecular graph datasets can only cover part of the overall molecular distribution, resulting in distribution shifts.
One data-centric approach to mitigating this problem is stable learning~\cite{stable}, which can be viewed as a special case of data sampling.
For example, SGL~\cite{SGL} introduces a stable graph learning framework, which can learn invariant patterns against selection bias in an unsupervised way.
DGNN~\cite{DGNN} re-weights the node weights to remove the spurious correlations in node representations.

\textbf{Heterophily.}
Most graph models rely on the homophily assumption, \ie, nodes belonging to the same class tend to connect with each other, to learn representations.
However, real-world graphs have mixed patterns, where heterophilic graphs exist, such as protein-protein interaction networks~\cite{Newman2002MixingPI}.
The homophilic ratio of graphs significantly affects the performance of graph models~\cite{FAGCN}.
Some methods use graph structure learning to find the non-local neighbors to alleviate the heterophily of graphs. For example,
Geom-GCN~\cite{Geom-GCN} constructs a latent geometric graph to enrich the potential homophilic neighbors.
AM-GCN~\cite{am-gcn} proposes to fuse the original graph and $k$NN graph of node features, which leverages the feature similarity to enhance the graph homophily.

\section{Future Directions}
\label{sec: future}

Data-centric AI is an emerging topic in deep learning. 
Here we present several promising future directions for data-centric graph learning.

\subsection{Standardized Graph Data Processing}

Existing graph construction and data processing methods rely on expert priors to define the structures, features, and labels, which seriously hinders the transferability of graph data across different domains.
For example, the graph models trained on citation networks cannot be used for social networks, due to their discrepancy in node features.
\textit{Is there a way that we can standardize the processing principles of graph data?}
If so, the ubiquitous graph data can be unified and the knowledge can be transferred across domains.
One possible approach is to use large language models (LLMs) to process graph data, unifying the node features in the language space~\cite{liu2023towards}.

\subsection{Continuous Learning of Graph Data}
Continuous learning aims to endow deep learning models with the ability to continuously learn new knowledge from a stream of data.
An interesting question is that \textit{can graph data also learn knowledge from the predictions of graph models?}
Graph models can learn semantic information from the raw graph data and remove some noise.
Based on the predictions of graph models, graph data can also be continuously optimized.
{\color{blue}}In this way, graph data can learn in conjunction with the graph model and adaptively adjust based on various graph models and downstream tasks, rather than heavily depending on prior knowledge or assumptions.
For example, graph condensation methods~\cite{jin2021graph, jin2022condensing} aim to use the gradients of graph models to generate new graph data, which can be seen as a special case of data continuous learning.

\subsection{Graph Data \& Models Co-development}

While we have frequently highlighted the importance of high-quality graph data for the success of model-centric graph learning, it is crucial to acknowledge the reciprocal relationship. It is foreseeable that optimal data manipulation strategies and model design mutually influence each other, and there is no single set of graph strategies/models that consistently performs best when paired with different graph models/strategies. Therefore, \textit{how to further encourage the collaborative development of graph data and models} is an important task for data-centric graph learning. The key to this lies in blurring the boundary between graph data and models, followed by the collaborative design of data-centric operations and model-centric methods. GraphStorm~\cite{zheng2023towards} adeptly focuses on both the development of graph data and the deployment of graph models, resulting in heightened effectiveness and efficiency.
\subsection{Graph Data for LLM-based Graph Models}
Motivated by the success of LLMs in language area, many works propose to explore graph models based on LLMs~\cite{liu2023towards}.
Similar to general LLMs, which require high-quality data for pre-training and whose capabilities largely depend on the pre-trained corpus and its preprocessing~\cite{zhao2023survey}, improvements in the quality and quantity of graph data are also key factors contributing to the effectiveness of graph models based on LLMs~\cite{liu2023towards}.
Despite the numerous data modification methods discussed above, most of these techniques are typically tailored for traditional GNNs. Therefore, there is a pressing need for further exploration into how to effectively modify graph data for LLM-based graph models.

\section{Conclusion}
In this survey, we give a comprehensive review of data-centric graph learning. We categorize existing methods from two perspectives: One is the learning stage, including pre-processing, training, and inference. Another is the data structure, including topology, feature, and label.
Through these two views, we carefully explain when to modify graph data and how to modify graph data to unlock the potential of the graph models.
Besides, we also introduce some potential issues of graph data and discuss how to solve them in a data-centric approach.
Finally, we propose several promising future directions in this field.
To sum up, we believe that data-centric AI is a viable path to general artificial intelligence, and data-centric graph learning will play an important role in graph data mining.

\section{ACKNOWLEDGEMENTS}
Writing a survey always involves more people than just the authors. We would like to express our sincere thanks to all those who worked with us on this paper. They are Boyu Chen, Haoran Dai, Ao Sun, Yue Yu, Yixin Xiao, Qi Zhang and Chunchen Wang. Additionally, this work was supported by the National Key Research and Development Program of China (No.2023YFC3303800).

\bibliographystyle{IEEEtran}
\bibliography{ref.bib}

\end{document}